\documentclass{article}

\usepackage{arxiv}

\usepackage[utf8]{inputenc} 
\usepackage[T1]{fontenc}    
\usepackage{hyperref}       
\usepackage{url}            
\usepackage{booktabs}       
\usepackage{amsfonts}       
\usepackage{nicefrac}       
\usepackage{microtype}      
\usepackage{lipsum}		
\usepackage{graphicx}
\usepackage{natbib}
\usepackage{doi}
\usepackage[most]{tcolorbox}
\usepackage{pmboxdraw}
\usepackage{dirtree}
\renewcommand{\DTcomment}[1]{%
  \hspace{0.7em}
  {\color{gray!30}\cleaders\hbox to 0.5em{\hss-\hss}\hfill}%
  \hspace{0.7em}
  {\color{gray}#1}%
}
\title{IJmond Industrial Smoke Segmentation Dataset}

\author{{Yen-Chia Hsu} \\
	Informatics Institute\\
	University of Amsterdam\\
	Netherlands \\
	\texttt{y.c.hsu@uva.nl} \\
	\And
	{Despoina Touska} \\
	Informatics Institute\\
	University of Amsterdam\\
	Netherlands \\
	\texttt{despoina.touska@student.uva.nl} \\
}

\renewcommand{\headeright}{Technical Report}
\renewcommand{\undertitle}{Technical Report}

\hypersetup{
pdftitle={IJmond Industrial Smoke Segmentation Dataset},
pdfsubject={Computer Vision, Image Segmentation, Industrial Smoke},
pdfauthor={Yen-Chia Hsu, Despoina Touska},
pdfkeywords={Computer Vision, Image Segmentation, Industrial Smoke},
}

\begin{document}
\maketitle

\begin{abstract}
	This report describes a dataset for industrial smoke segmentation, published on a figshare repository -- \url{https://doi.org/10.21942/uva.31847188}. The dataset is licensed under CC BY 4.0\footnote{{CC BY 4.0 -- \url{https://creativecommons.org/licenses/by/4.0/}}}.
\end{abstract}


\section{Introduction}

The dataset contains raw and cropped images for industrial smoke segmentation. The task is to identify the pixels in the image that belongs to a part of the industrial smoke emission. The standard way for using this dataset is to cast the task as pixel-level classification. However, one can also go for the polygon-based image segmentation approach. We have in total 900 raw images with 1209 polygons. Among them, 7 raw images have no smoke. Also, we have in total 2074 cropped images, and among them, 1109 images have smoke (and have pixel-level masks).

\section{Data Source}

\begin{figure}[h]
	\centering
	\includegraphics[width=0.7\textwidth]{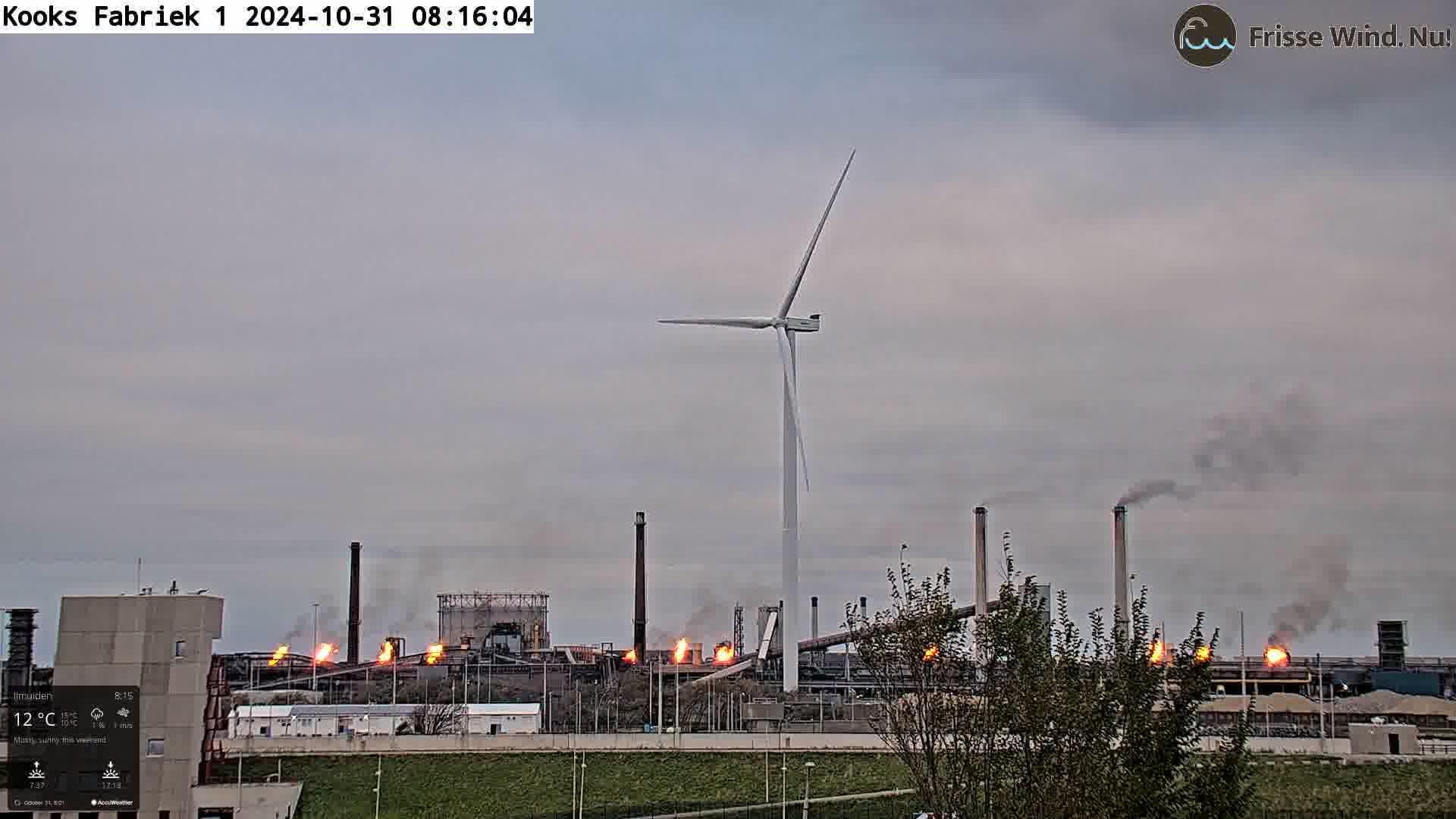}
    \includegraphics[width=0.7\textwidth]{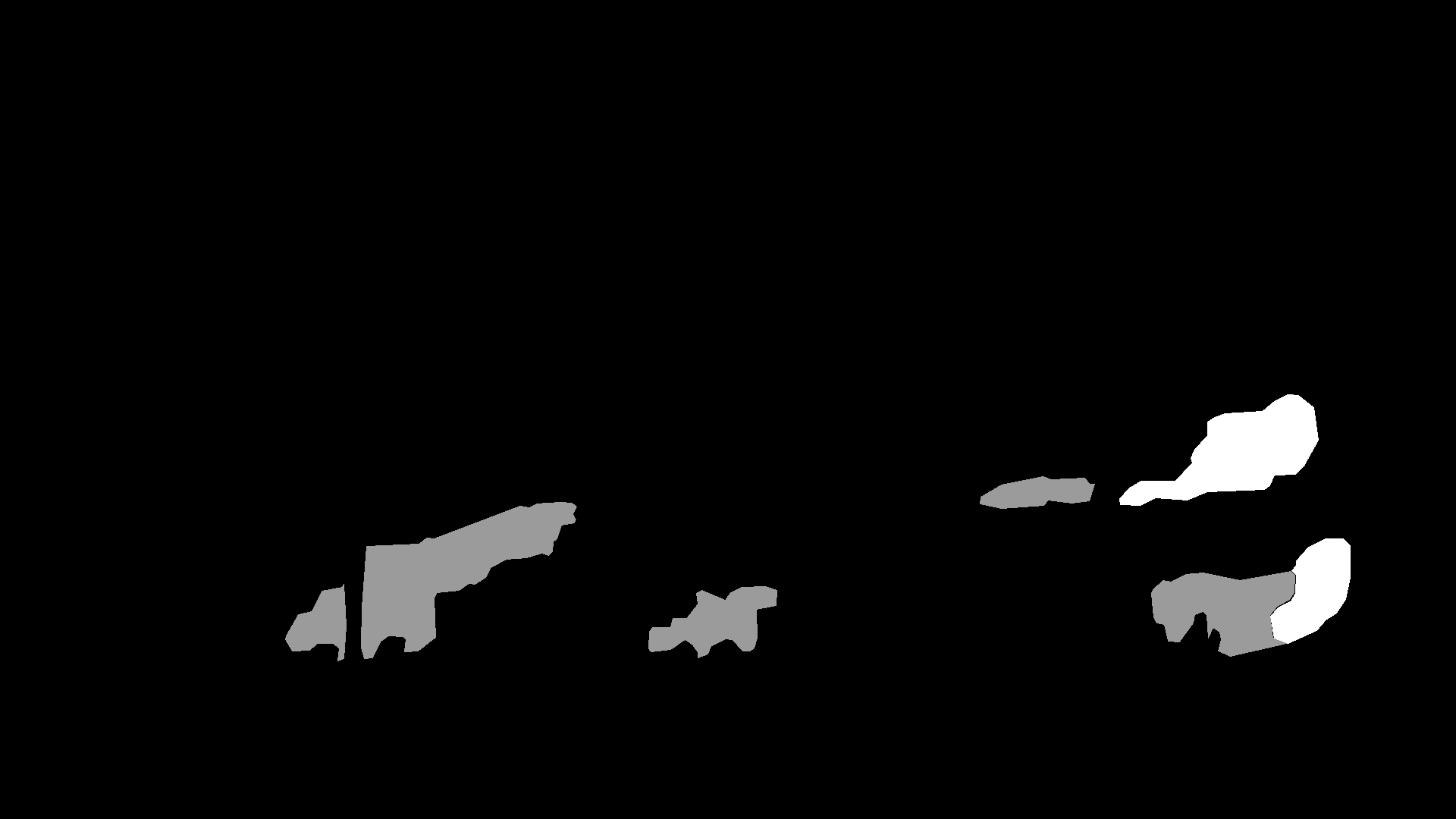}
	\caption{Example of a raw image that is scrapped from the "kooks\_1" monitoring camera, as well as its annotated mask. White pixels (with value 255) indicate high opacity smoke, and gray pixels (with value 155) indicate low opacity smoke. Background pixels have black color (value 0).}
	\label{fig:fig1}
\end{figure}

\begin{figure}[h]
	\centering
	\includegraphics[width=0.7\textwidth]{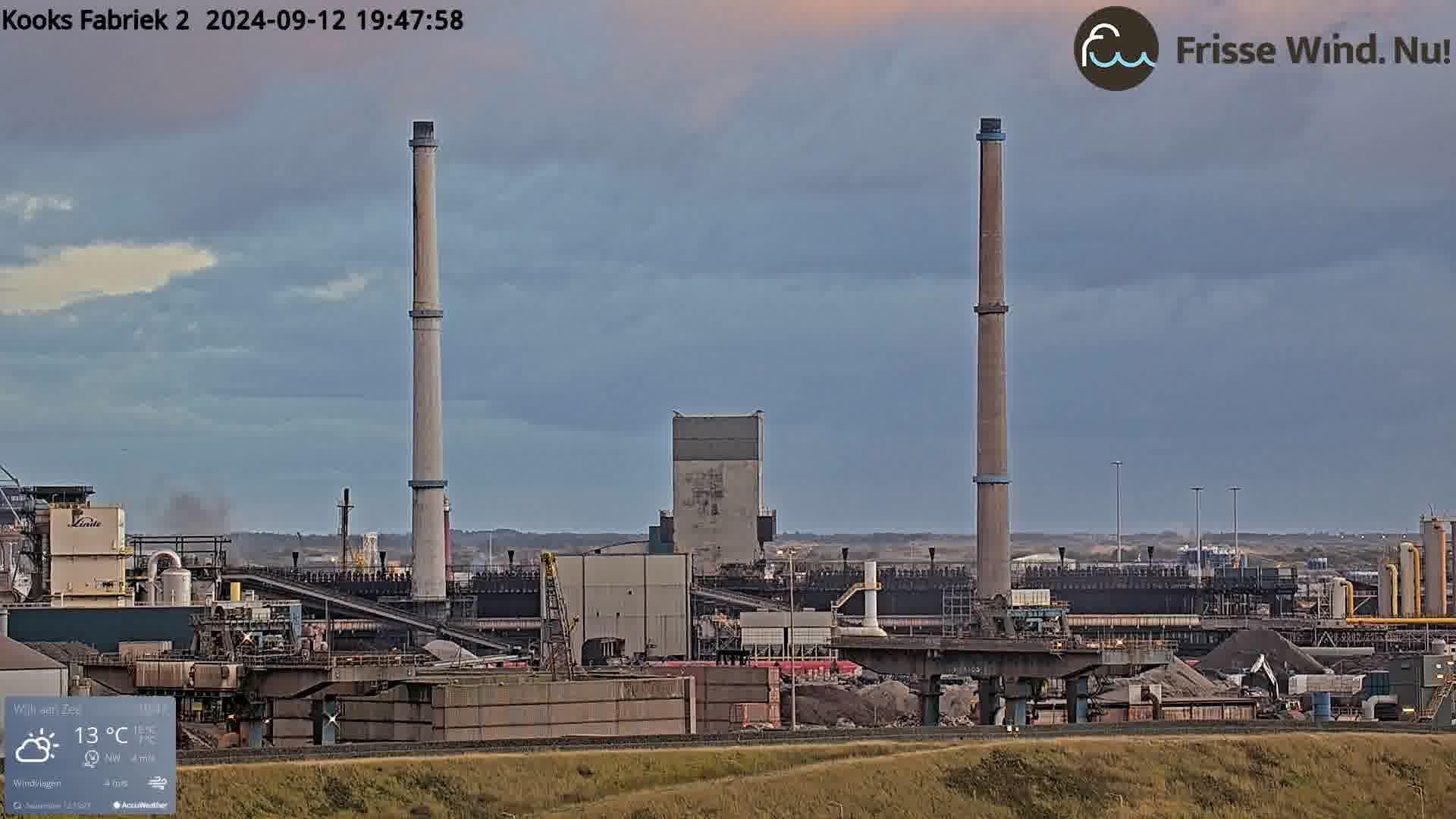}
    \includegraphics[width=0.7\textwidth]{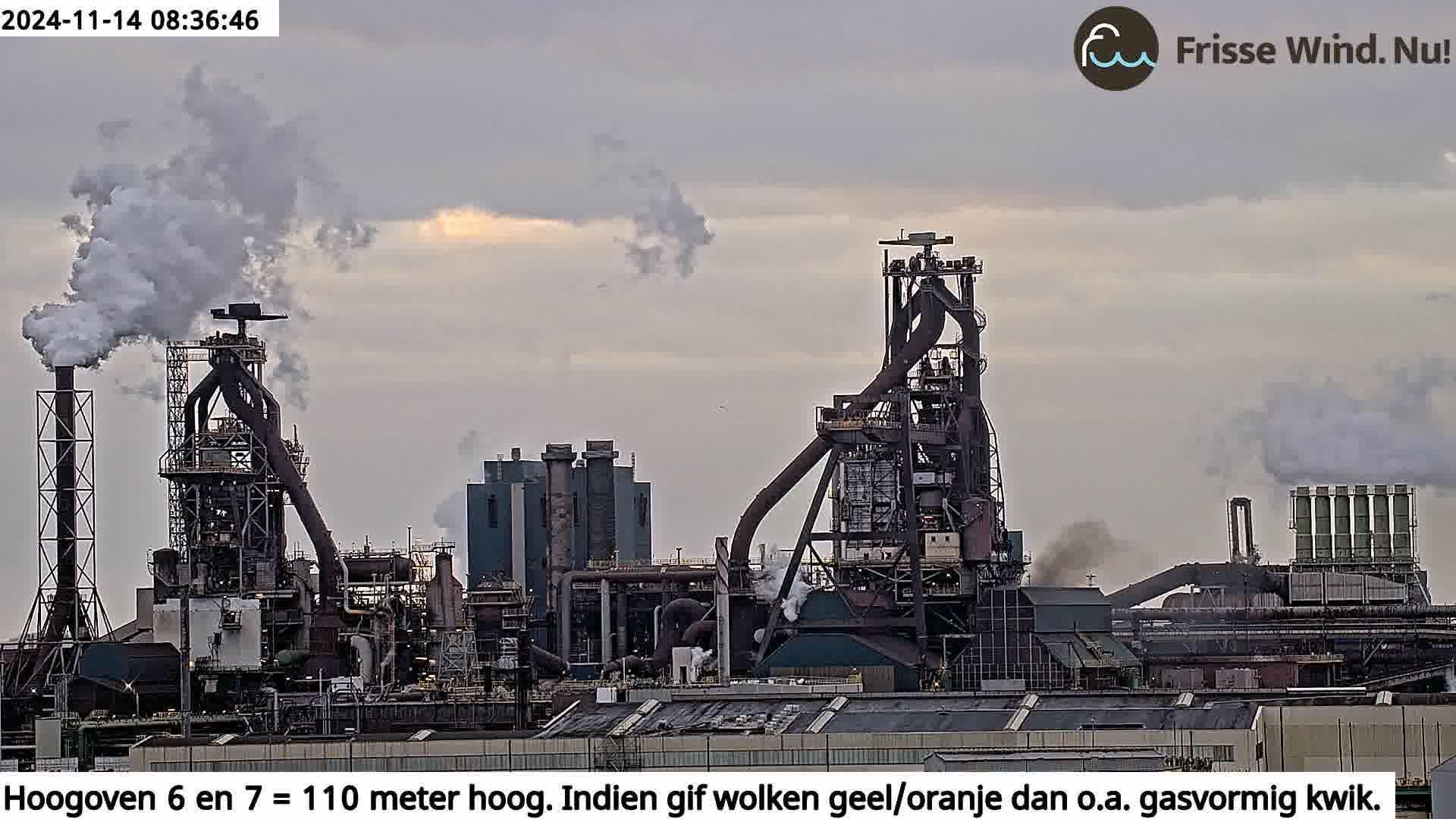}
	\caption{Examples of raw images from "kooks\_2", and "hoogovens\_6\_7".}
	\label{fig:fig3}
\end{figure}

The images come from three stationary cameras that are operated by FrisseWind.nu \footnote{FrisseWind.nu -- \url{https://www.frissewind.nu/}} to monitor local air pollution in the IJmond region in the Netherlands. FrisseWind.nu is a local organization that is dedicated to improving local air quality. We call the three cameras "kooks\_1", "kooks\_2", and "hoogovens\_6\_7" based on the facilities that they monitor, which are reflected in the prefix of file names in the dataset. We scrapped the images directly from the cameras under the consent from FrisseWind.nu. See Figure~\ref{fig:fig1} for an example of the raw image for the "kooks\_1" camera, and see Figure~\ref{fig:fig3} for examples of the raw images for the "kooks\_2" and "hoogovens\_6\_7" cameras.

\section{Annotation}

We annotated the data using the Roboflow\footnote{Roboflow -- \url{https://roboflow.com/}} platform. The annotation was done using the smart polygon tool on Roboflow, where points were first used as prompts for the Segment Anything model~\citep{kirillov2023segment} to generate initial masks, and then the polygons were manually refined by the first and second authors..

The first and second authors labeled the data jointly. The first author has experiences working with industrial smoke detection and has received training in smoke reading using EPA Method 9\footnote{EPA Method 9 -- \url{https://www.epa.gov/emc/method-9-visual-opacity}}. All the segmentation masks were manually checked and edited by the first author to ensure quality.

We annotated two types of masks: low opacity (corresponding to less than 50\% opacity according to EPA Method 9) and high opacity (more than 50\% opacity). However, visual opacity is know to be affected by lighting conditions, which depends on the position of the sun. In order to correctly determine the visual opacity, the sun needs to be in a specific range of positions relative to the camera, which is not always the case in our dataset. Under these circumstances, the first author tried to still determine the visual opacity using the original rule in EPA Method 9. But please notice that the visual opacity may be different in reality, especially in the situation when the opacity is near the boundary when deciding the class labels (i.e., 50\% visual opacity).

See Figure~\ref{fig:fig1} for an example of the annotated mask. The mask is an image in the uint8 (8-bit unsigned integer) format. White pixels (with value 255) indicate high opacity smoke, and gray pixels (with value 155) indicate low opacity smoke. Background pixels have black color (value 0).

\section{Cropping and Splitting}
\label{sec:crop_split}

\begin{figure}[h]
	\centering
	\includegraphics[width=0.22\textwidth]{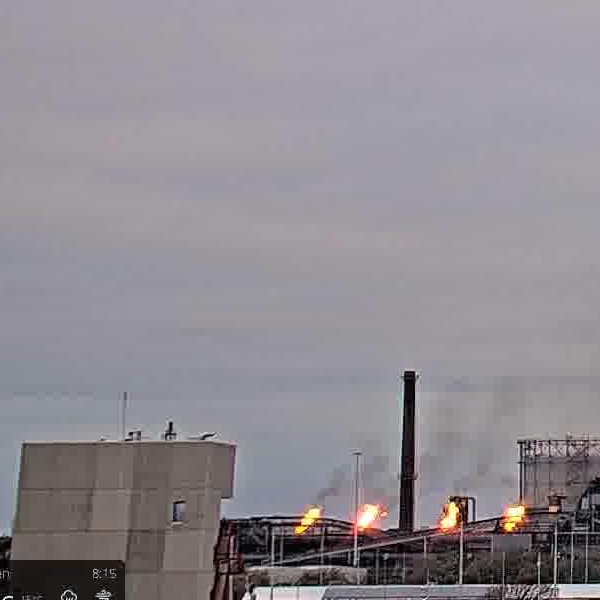}
    \hspace{1mm}
    \vspace{2mm}
    \includegraphics[width=0.22\textwidth]{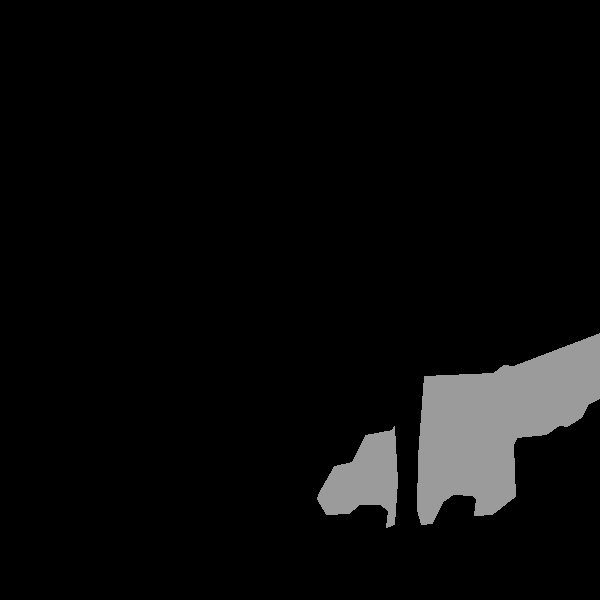}
    \hspace{4mm}
    \vspace{2mm}
    \includegraphics[width=0.22\textwidth]{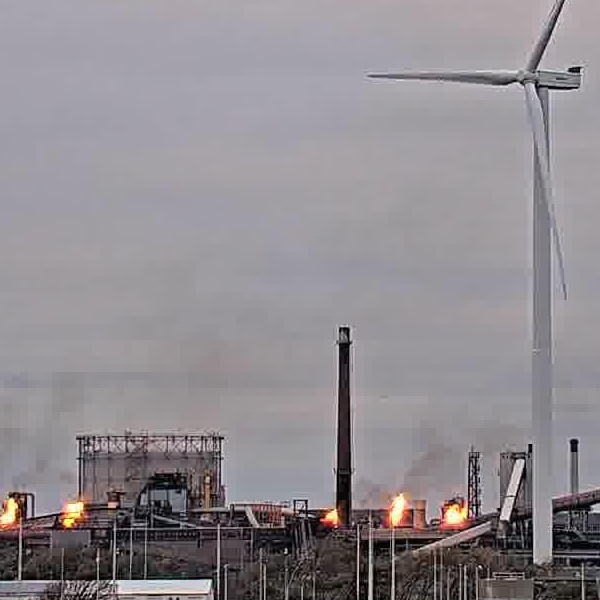}
    \hspace{1mm}
    \vspace{2mm}
    \includegraphics[width=0.22\textwidth]{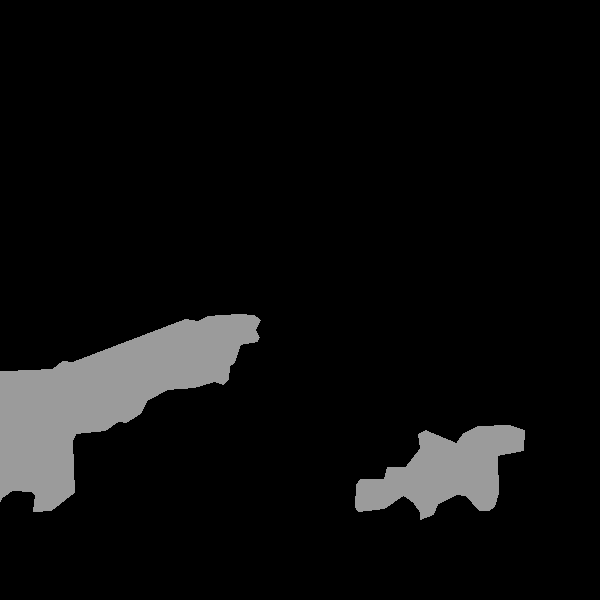}
    \vspace{2mm}
    \includegraphics[width=0.22\textwidth]{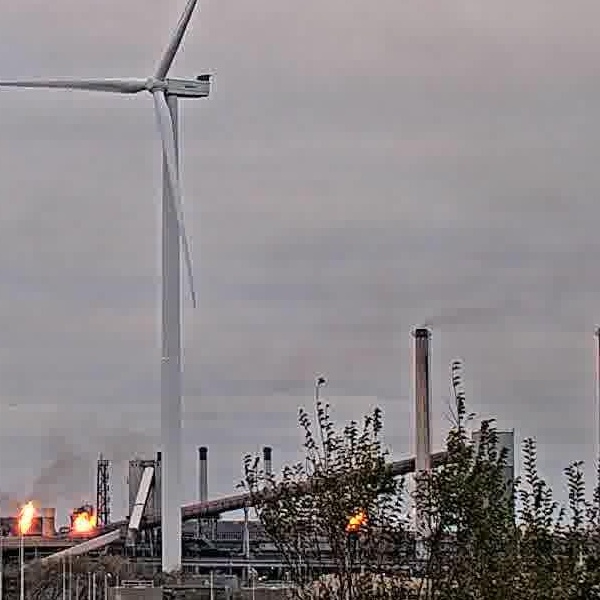}
    \hspace{1mm}
    \includegraphics[width=0.22\textwidth]{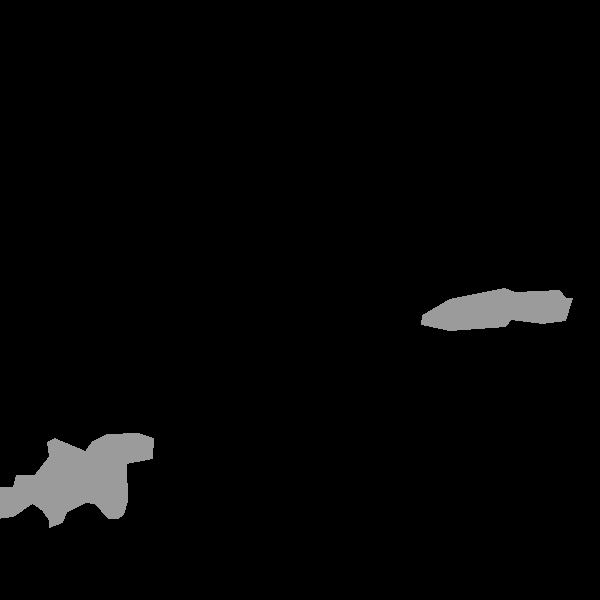}
    \hspace{4mm}
    \includegraphics[width=0.22\textwidth]{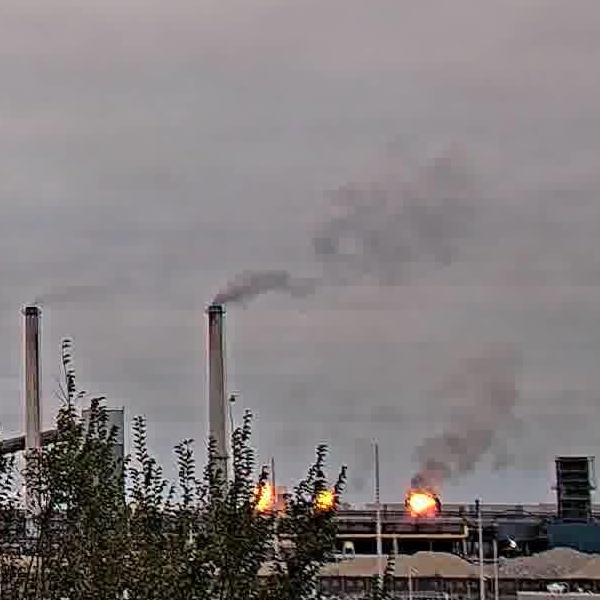}
    \hspace{1mm}
    \includegraphics[width=0.22\textwidth]{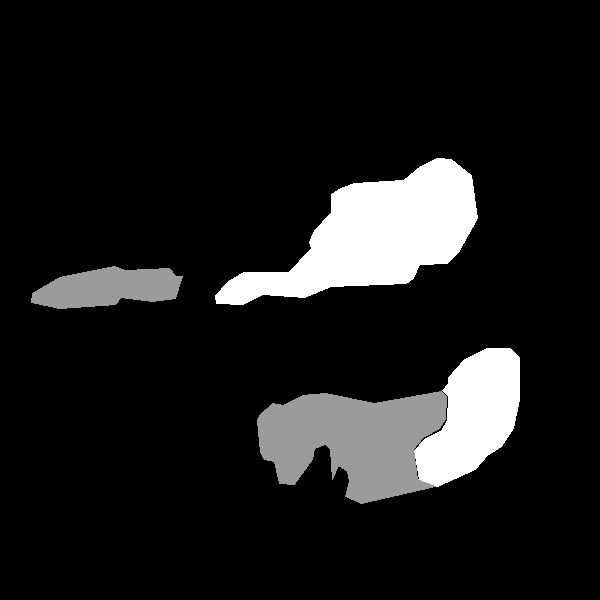}
	\caption{Examples of cropped images from the raw image in Figure~\ref{fig:fig1} and their corresponding masks.}
	\label{fig:fig2}
\end{figure}

After annotation, we cropped the raw images into smaller ones, and then we split the cropped images into training, validation, and test sets. The cropping is based on the logic in Project RISE~\cite{hsu2021project} using domain knowledge about where smoke emissions frequently occur. Figure~\ref{fig:fig2} shows the cropped images from the raw image in Figure~\ref{fig:fig1}.

We provide two different types of splits: by timestamp, and by camera. For the timestamp split, the training set contains the first 70\% of data sorted by timestamps, the validation set contains the next 10\%, and the test set contains the final 20\%. For the camera split, the training set contains the first 80\% of the "kooks\_2" camera sorted by timestamps, the validation set contains the rest of 20\% of "kooks\_2" sorted by timestamps, and the test set contains all images from the "hoogovens\_6\_7" and "kooks\_1" cameras. You can find the range of dates, camera names, and number of images for each split in the "metadata.json" files (see Section~\ref{sec:fs} for the explanation of its format).

For each split, we separate them into two parts: with and without smoke emissions. We call them "with mask" or "without mask" in the file names. But it is important to know that they do \textbf{NOT} mean labeled/unlabeled data. The reason for doing this is to add more flexibility when training models, which means researchers can choose to use only the images with masks (i.e., positive images, with smoke emissions) or a combination of them, such as using images with masks together with a small sampled set from the images without masks (i.e., negative images, without smoke).

Specifically for the training set, we create five different versions of it, with 100\%, 80\%, 60\%, 40\%, and 20\% of the training data, sorted by timestamps. The 100\% training split is the full training set. The 80\% training split is the last 80\% of the training set sorted by timestamps, and so on. These percentages are the "last" part in the training set according to sorted timestamps to ensure that the timestamps, when considered together with the validation and test sets, are continuous. The reason for this percentage splitting is to simulate different amounts of available data for training.

\section{File Structure}
\label{sec:fs}

The file structure is explained below. All the images are placed under the "test" folder. Inside the folder, there are original raw images, masks for the raw images, the polygon annotations in the MS COCO format, and the cropped images/masks. Inside the "cropped" folder, there are cropped images, cropped masks, splits (by timestamps and by cameras), a metadata file (with information for each cropped image), and two text files that contain the image-mask pairs (one with mask, and one without mask). For the images without masks, the masks files (.png) are just images with black color (this is just for convenience so that the model can just read the empty masks).

\begin{tcolorbox}[title=General file structure, colupper=black, colback=gray!5, colframe=blue!50!black, fontupper=\ttfamily]
\dirtree{%
.1 ijmond\_seg \DTcomment{the root folder}.
.2 README.dataset.txt \DTcomment{general dataset information}.
.2 README.roboflow.txt \DTcomment{dataset version information}.
.2 test \DTcomment{the folder that contains the data}.
.3 cropped \DTcomment{all cropped images and masks}.
.4 images \DTcomment{all cropped camera images}.
.5 XXX.jpg.
.5 \dots.
.4 masks \DTcomment{all cropped masks}.
.5 XXX.png.
.5 \dots.
.4 splits \DTcomment{a folder that contains the splits}.
.4 test\_with\_mask.txt \DTcomment{paths for image-mask pairs (with masks)}.
.4 test\_without\_mask.txt \DTcomment{paths for image-mask pairs (with no masks)}.
.4 metadata.json \DTcomment{metadata for each cropped image}.
.3 images \DTcomment{all raw camera images}.
.4 XXX.jpg.
.4 \dots.
.3 masks \DTcomment{all masks for raw camera images}.
.4 XXX.png.
.4 \dots.
.3 \_annotations.coco.json \DTcomment{polygon annotations in the MS COCO format}.
}
\end{tcolorbox}

The metadata file has the following format. Please ignore the "cropped\_npy\_name" and "cropped\_mask\_npy\_name" fields. The "x" and "y" fields mean the coordinate of the top-left corner of the cropped image that is relative to the raw image. In other words, if we place the top-left corner of the cropped image in the original position of the raw image, the coordinate would be (x, y). It is important to know that the origin (i.e., when x and y are both 0) of the coordinate system for both the original and cropped images is at the top-left corner.

\begin{tcolorbox}[title=Format of "metadata.json" under the cropped folder, colupper=black, colback=gray!5, colframe=blue!50!black, fontupper=\ttfamily]
\begin{verbatim}
[
  {
    "original_file_name": "XXX.jpg",
    "cropped_file_name": "XXX_crop_1.jpg",
    "original_mask_name": "XXX.png",
    "cropped_mask_name": "XXX_crop_1.png",
    "cropped_npy_name": "XXX.npy",
    "cropped_mask_npy_name": "XXX_crop_1.npy",
    "x": 128,
    "y": 54,
    "width": 900,
    "height": 900,
    "view_id": 1,
    "camera_name": "kooks_2"
  },
  ...
]
\end{verbatim}
\end{tcolorbox}

Next, we explain the file structure inside the "splits" folder. See Section~\ref{sec:crop_split} to understand the logic of splitting and cropping. Inside the "splits" folder, we have two sub-folders: one for the splits based on timestamps, and one based on cameras. The text files are image-mask pairs (the same as mentioned before). There is a "metadata.json" file for each split type, documenting the coverage of cameras, date ranges, and number of images.

\begin{tcolorbox}[title=File structure inside the "splits" folder, colupper=black, colback=gray!5, colframe=blue!50!black, fontupper=\ttfamily]
\dirtree{%
.1 splits.
.2 split\_by\_timestamp \DTcomment{the split based on timestamps}.
.3 train \DTcomment{training set}.
.4 100\_with\_masks.txt \DTcomment{100\% of the training set with smoke}.
.4 100\_without\_masks.txt.
.4 80\_with\_masks.txt \DTcomment{80\% of the training set with smoke}.
.4 80\_without\_masks.txt.
.4 60\_with\_masks.txt \DTcomment{60\% of the training set with smoke}.
.4 60\_without\_masks.txt.
.4 40\_with\_masks.txt \DTcomment{40\% of the training set with smoke}.
.4 40\_without\_masks.txt.
.4 20\_with\_masks.txt \DTcomment{20\% of the training set with smoke}.
.4 20\_without\_masks.txt.
.3 val\_with\_masks.txt \DTcomment{validation set with smoke}.
.3 val\_without\_masks.txt.
.3 test\_with\_masks.txt \DTcomment{test set with smoke}.
.3 test\_without\_masks.txt.
.3 metadata.json \DTcomment{coverage of cameras and dates for each txt file}.
.2 split\_by\_camera \DTcomment{the split based on cameras}.
.3 \dots \DTcomment{same file structure as in the "split\_by\_timestamp" folder}.
}
\end{tcolorbox}

\section{Acknowledgment}

This work is supported by the Dutch Research Council (NWO, Dutch: Nederlandse Organisatie voor Wetenschappelijk Onderzoek) under grant number OSF23.2.036 (Open Science Fund). We thank FrisseWind.nu in giving us permissions to use the camera images for research. We also thank the Roboflow platform for providing a free plan for data annotation.

\bibliographystyle{unsrtnat}
\bibliography{references}  






\end{document}